\documentclass{sig-alternate-05-2015}

\begin{document}
	
	
	\makeatletter
	\def\@copyrightspace{\relax}
	\makeatother
	
	\title{3D Binary Signatures}

	\numberofauthors{2} 
	%
	\author{
		%
		%
		\alignauthor
		Siddharth Srivastava\\
		\affaddr{Department of Electrical Engineering}\\
		\affaddr{Indian Institute of Technology, Delhi, India}\\
		\email{eez127506@ee.iitd.ac.in}
		\alignauthor
		Brejesh Lall\\
		\affaddr{Department of Electrical Engineering}\\
		\affaddr{Indian Institute of Technology, Delhi, India}\\
		\email{brejesh@ee.iitd.ac.in}\\
	}
	
	\maketitle
	\begin{abstract}
		In this paper, we propose a novel binary descriptor for 3D point clouds. The proposed descriptor termed as 3D Binary Signature (3DBS) is motivated from the matching efficiency of the binary descriptors for 2D images. 3DBS describes keypoints from point clouds with a binary vector resulting in extremely fast matching. The method uses keypoints from standard keypoint detectors. The descriptor is built by constructing a Local Reference Frame and aligning a local surface patch accordingly. The local surface patch constitutes of identifying nearest neighbours based upon an angular constraint among them. The points are ordered with respect to the distance from the keypoints. The normals of the ordered pairs of these keypoints are projected on the axes and the relative magnitude is used to assign a binary digit. The vector thus constituted is used as a signature for representing the keypoints. The matching is done by using hamming distance. We show that 3DBS outperforms state of the art descriptors on various evaluation metrics.
	\end{abstract}

	%
	%

	\ccsdesc[500]{Computing methodologies~Interest point and salient region detections}
	\ccsdesc[500]{Computing methodologies~Matching}
	\ccsdesc[500]{Computing methodologies~Computer graphics}
	\ccsdesc[300]{Computing methodologies~Object detection}
	\ccsdesc[300]{Computing methodologies~Object recognition}

	%
	%
	
	%
	%
	\printccsdesc
	
	
	\keywords{3D Descriptors; 3D Matching; 3D Object Recognition}
	\section{Introduction}
	Humans visualize a three dimensional world. Therefore, the ability to obtain and process information especially from visual senses in three dimensions has always been an exciting and potentially promising area of research. The success of various feature extraction and classification related tasks in 2D image analysis can be attributed to the availability of large scale annotated datasets such as ImageNet \cite{deng2009imagenet,russakovsky2015imagenet}. With the growing availability of 3D scanning technologies such as Kinect \cite{zhang2012microsoft}, LIDAR etc., the availability of large and quality 3D datasets is also increasing. This has opened many research areas from 2D image analysis such as Object Recognition, Object Retrieval, Object Classification, Segmentation etc. for three dimensional data as well. 
	
	Local features and Deep Learning based approaches especially Convolutional Neural Networks (CNNs) have been successfully used for solving many research problems involving 2D images \cite{sanchez2013image, chan2015pcanet,krizhevsky2012imagenet,srivastava2015characterizing, jiang2015revisiting}. Similar to 2D domain, various local feature  and deep learning based models have been proposed for 3D point clouds. Although, CNN based techniques have been successfully proposed for 3D object classification \cite{socher2012convolutional, gupta2014learning} and recognition \cite{maturana2015voxnet}, but 3D local features are state of the art in many tasks such as 3D object recognition and classification \cite{guo20143d, salti2014shot}, 3D scene reconstruction \cite{guo2014accurate}, 3D model retrieval \cite{gao2014view} etc. Comparisons of descriptors on various benchmarks \cite{guo2016comprehensive} show that high descriptiveness with low storage requirements and reasonable computational complexity depending upon the number of points in the point cloud constitute key requirements for a descriptor. Moreover, for practical applications involving 3D object matching, recognition, classification and reconstruction, the efficiency in feature matching and lower storage requirements become important \cite{alexandre20123d}. In order to address these issues, we propose a novel 3D binary descriptor deriving motivation from binary descriptors for 2D images \cite{calonder2012brief, leutenegger2011brisk, rublee2011orb}. These descriptors have reported to match or outperform traditional SIFT like descriptors \cite{lowe1999object, bay2006surf} with significantly lower computational complexity and storage requirements. The proposed technique encodes the differences in the projection of normals into binary vectors. The projection is computed for nearest neighbours of a keypoint and aligned with a local reference frame. The nearest neighbours are chosen based on their angular orientation on a 2D projection plane. The binary descriptor thus generated is matched using Hamming Distance.
	We show that the proposed binary descriptor outperforms the state of the art on various benchmarks. 
	
	The closest work to ours is B-SHOT \cite{prakhya2015b} which generates a binary vector from the popular Signature of Histograms of Orientations (SHOT) \cite{salti2014shot} descriptor. It quantizes the real valued SHOT descriptor to a binary vector. The major difference in the proposed technique is that the binary vector is generated directly from the point cloud data  instead of first computing a real valued vector and quantizing it which results in an additional overhead. For highlighting the novelty of the technique, we refer to \cite{guo2016comprehensive} where authors categorize 3d descriptors into two classes based on the methodology of generating histograms. First class of descriptors computes histograms either by computing a local reference frame and  accordingly divide the support region spatially. Various spatial distribution measurements are then accumulated into histograms. The second class of descriptors, encodes  geometric attributes such as normals, principal curvatures etc. of the points on the surface in the local neighbourhood the keypoint. The proposed approach is a hybrid of these two approaches. We define a local surface with the nearest neighbours of a keypoint with an angular constraint. Then similar to the first class we use a Local Reference Frame (LRF) for aligning this local surface followed by computing normals and encoding the projective difference among them which is similar to the second class of descriptors. Since the proposed technique does not require computation of histograms, therefore to highlight this difference, we term the extracted descriptors as \textit{signatures}. 
	In view of the above discussion, the main contributions of this paper are: 
	\begin{itemize}
		\item We propose a hybrid approach for constructing a highly distinctive yet compact 3D binary descriptor based on encoding differences among normal projections among nearest neighbours of a keypoint. To the best of our knowledge, this is the first work to directly generate 3D descriptors from  point cloud data. 
		\item We show that the proposed 3D Binary Signature (3DBS) outperforms state of the art descriptors on common evaluation benchmarks. 
	\end{itemize}
	
	The paper is organized as follows: In Section \ref{relatedworks}, we discuss the techniques that are similar to ours and have motivated the construction of the proposed 3D Binary Signature. In Section \ref{methodology}, we detail the generation of 3D Binary Signatures (3DBS) followed by experimental results in Section \ref{results}. Finally, Section \ref{conclusion} concludes this paper. 
	
	\section{Related Work}\label{relatedworks}
	Numerous 3D descriptors have been proposed in the past two decades. Broadly these descriptors fall into two categories. First, the descriptors that encode the geometric attributes of a spatial region as histograms. Second, the descriptors that encode as histograms the statistical properties of a region.  Another way of looking at these descriptors is whether a Local Reference Frame (LRF) has been used for forming the descriptor. The descriptors without LRF utilize local statistics such as normals etc. to form the descriptors while those using LRF encode information from spatial distribution or geometric properties of neighbouring points to form the descriptors. Authors in \cite{guo2016comprehensive} provide an exhaustive evaluation of various 3D binary descriptors. In the rest of this section, we discuss a few techniques which are either (i) similar to the proposed technique or (ii) use surface normals for generating the descriptors. We first discuss a few techniques which particularly focus on speed and memory efficiency, followed by robust and general purpose techniques. 
	
	B-SHOT \cite{prakhya2015b} is a 3D binary feature descriptor based on the popular SHOT descriptor. It is generated by quantizing the $352$ length real valued SHOT descriptor to $352$ length binary vector.  The quantization begins by dividing the SHOT descriptor into sets of four values. The relative magnitudes of various combinations of values in each set are compared to assign a corresponding four-bit binary vector.  There is a loss of information due to this quantization but it is compensated by significant gains in the descriptor matching efficiency and storage requirement. Signature of Histograms of Orientations (SHOT) \cite{salti2014shot}, which is the basis behind B-SHOT, is based upon encoding into histograms, the surface normals in a spatial distribution. It works by constructing an LRF for a keypoint and the points in the support region are aligned with this LRF. The support region is then divided into several volumes. Each volume results in a local histogram by accumulation of point counts into bins as per the angles between normal of points in the support region and the keypoint. The final descriptor is computed by concatenating all the local histograms.  Spin Image (SI) \cite{johnson1999using} descriptor uses the surface normal at a keypoint as Local Reference Axis. Then in-plane and out-plane distances are computed for each point in the support region which is discretized into a 2D array. The final histogram is generated by binning the points from the support region to the 2D array constructed earlier. The dimension of the SI descriptor is equal to the number of bins across each dimension of the in-plane and out-plane space. Rotational Projection Statistics (RoPS) \cite{guo2013rotational} constructs an LRF for each keypoint aligning it with the local surface to achieve invariance to transformations. The points on the local surface are rotated around $x$, $y$, and $z$ axis and the corresponding support region are further projected onto the coordinate planes ($xy$, $xz$ and $yz$). The planes are then divided into several bins and the number of points in each bin is counted.  Various statistics are calculated on these bins and they are concatenated to form the final descriptor. 3D Shape Context (3DSC) \cite{frome2004recognizing} uses normal as Local Reference Axis. Further, it divides the support region into many bins along azimuth, elevation and radial dimensions. The descriptor is generated by weighted accumulation of the points lying in each bin. Unique Shape Context (USC) \cite{tombari2010unique} improves the memory requirement and computational efficiency of 3DSC by avoiding computation of multiple descriptors at a keypoint. This is achieved by defining an LRF (same as that used in SHOT) at a keypoint  and aligns the local surface accordingly. The support region is then divided into bins and the weighted accumulation is performed for the points lying in corresponding bins. 
	
	\section{Proposed Methodology}\label{methodology}
\begin{figure*}[h]\label{fig:flowdiag}
		\caption{Construction of 3D Binary Signature}
		\centering
		\includegraphics[width=0.5\textwidth]{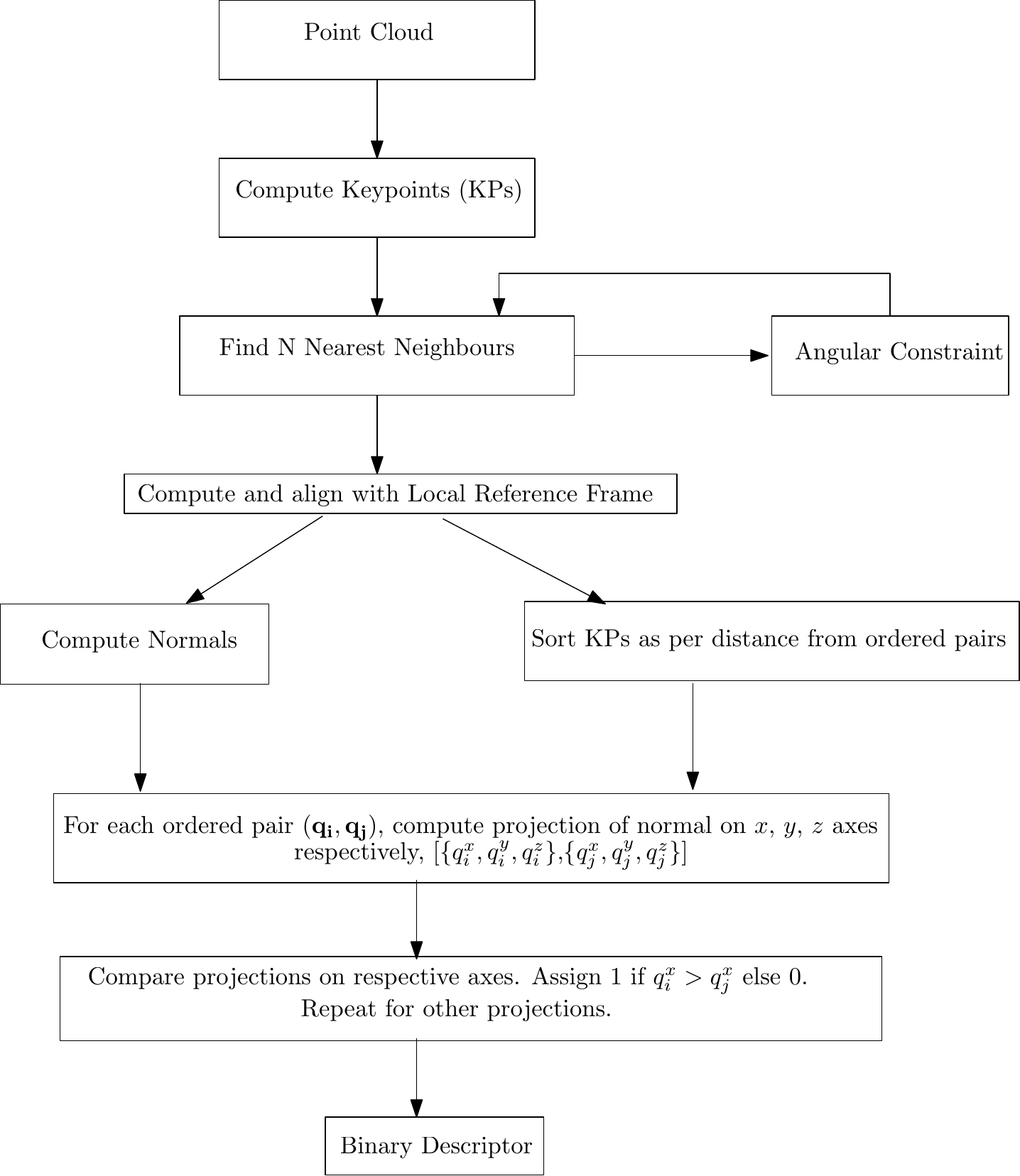}
	\end{figure*}

	In this section, we discuss the methodology for generating the 3D Binary Signatures. Figure \ref{fig:flowdiag} shows the construction pipeline of the proposed binary descriptor. 
	
	The proposed methodology is abstractly inspired from the descriptor construction pipeline of 2D binary descriptors. Generation of 2D binary descriptors in general involves three steps: a) Choosing a sampling pattern b) Orientation Compensation and c) Selection of Sampling pairs. The sampling pattern in the current methodology is obtained with the help of nearest neighbours while Local Reference Frame compensates for invariance to orientation and other transformation. The sampling pairs are the ordered nearest neighbour pairs which are compared for encoding the difference in projections of the normals to a binary vector. We describe these steps in detail in the following subsections.
	
	\begin{figure*}[h]\label{fig:method1}
		\caption{Visualization of a) Angular Constraint b) Comparison of projection of Normals}
		\centering
		\includegraphics[width=0.75\textwidth]{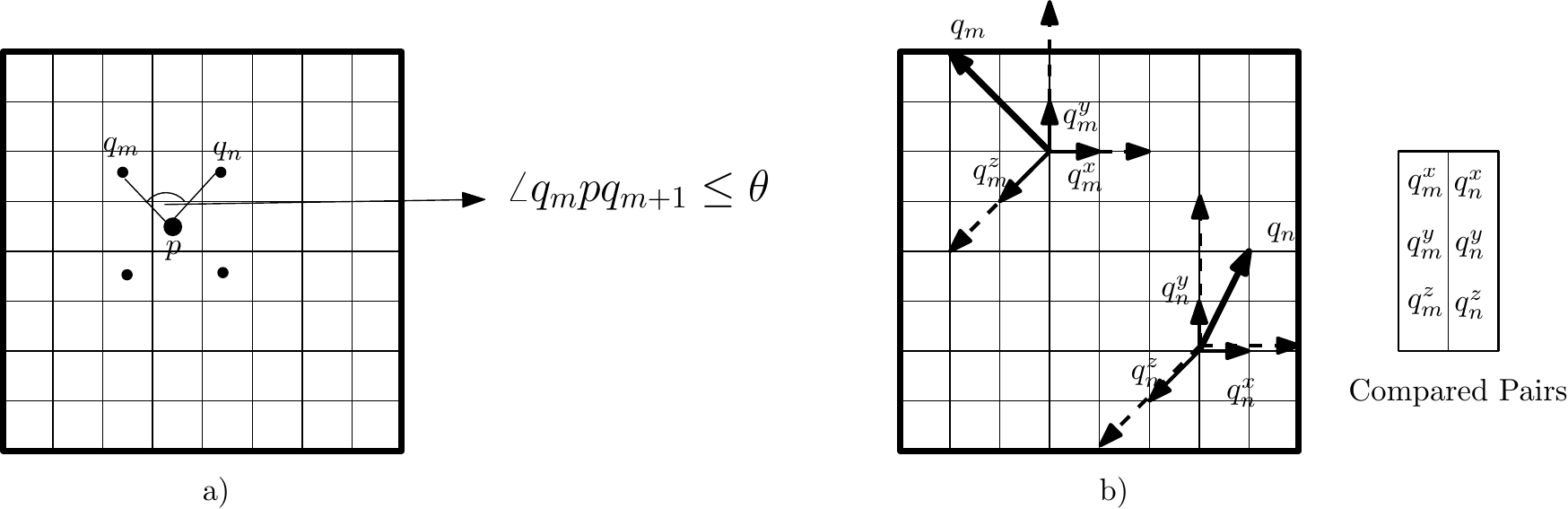}
	\end{figure*}
	
	\subsection{Keypoint Detection}\label{sec:kpdet}
	The keypoints can be detected using any of the standard keypoint detection techniques. From our experiments on various keypoint detectors, namely, Intrinsic Shape Signature (ISS) \cite{zhang2012microsoft}, MeshDoG \cite{zaharescu2009surface}, Keypoint Quality (KPQ) \cite{mian2010repeatability}, Harris3D \cite{sipiran2011harris}, Heat Kernel Signatures \cite{sun2009concise} and 3D SIFT \cite{flint2007thrift}, it was observed that ISS and Harris3D performed  best in our experiments. Authors in \cite{tombari2013performance} present a comprehensive survey on various keypoint detectors. They show that Intrinsic Shape Signatures(ISS) \cite{zhong2009intrinsic} demonstrate the highest repeatability while being the most efficient keypoint detector. Moreover, the work in \cite{alexandre20123d} showed that selection of an appropriate keypoint detector has a reasonable impact on the performance of a descriptor. Therefore, for further experiments in this work, ISS has been chosen as the keypoint detector.
	
	\subsection{Nearest Neighbours with Angular Constraint} \label{subsec:nn}
	The objective of this criteria is to uniformly distribute the nearest neighbours along the surface of the point cloud. The process is pictorially shown in Figure 2(a). Instead of a spherical support region, we define the local surface for each keypoint with its $N$ nearest neighbours. Nearest neighbours are traditionally computed based on a distance criterion. Authors in \cite{linsen2001point} observe that computing nearest neighbour in such a manner may not be an optimal choice for local surface representation and hence introduce an angle criterion. We therefore use the angle criterion to distribute neighbours around the keypoint $\boldsymbol{p}$, by projecting the neighbours of the points on a plane $\boldsymbol{\phi}$ which is  best fitting plane of neighbours of $\boldsymbol{p}$. The plane $\boldsymbol{\phi}$ is obtained as a least squares best fitting plane which turns the approximation into a quadratic form and therefore can be solved efficiently. The points thus obtained are sorted as per distance and for each neighbour of $\boldsymbol{p}$, if the angle between $\boldsymbol{p}$, its neighbour $\boldsymbol{q_m}$ and its successor $\boldsymbol{q_{m+1}}$, does not exceed a threshold $\theta$ i.e. $\angle \boldsymbol{q_mpq_{m+1}} \leq \theta$, it is considered added to the set of nearest neighbours. This process is repeated till $N$ neighbours have been found. Angle criterion also helps in avoiding ambiguity in the order of neighbours having the same distance from a keypoint. For instance, if $\boldsymbol{q_1}, \boldsymbol{q_2}$ and $\boldsymbol{q_3}$ are neighbours of keypoint $\boldsymbol{p}$ with the following distance relation, $||\boldsymbol{p}-\boldsymbol{q_2}||_2 = ||\boldsymbol{p}- \boldsymbol{q_3}||_2 < ||\boldsymbol{p}-\boldsymbol{q_1}||_2$. If $\angle{\boldsymbol{q_1pq_2}} < \angle{\boldsymbol{ q_1pq_3}}$, $\boldsymbol{q_2}$ is listed before $\boldsymbol{q_3}$ avoiding any ambiguity. 
	
	\subsection{Alignment with Local Reference Frame}
	To infuse invariance to various transformations such as translation and rotation along with robustness to noise and clutter, the local surface formed previously is aligned with a Local Reference Frame. Authors in \cite{petrelli2011repeatability} show that repeatability of an LRF has a direct impact on the robustness and descriptive ability of a descriptor. Therefore, we construct an LRF using the technique of \cite{salti2014shot} which is the basis for the popular SHOT descriptor. It computes a weighted covariance matrix $\boldsymbol{M}$ given by Eq. \ref{eq:covmatrix} around the keypoint $\boldsymbol{p}$ where the distant points are assigned smaller weights. 
	
	\begin{equation} \label{eq:covmatrix}
		\boldsymbol{M} = \frac{1}{\sum_{i:d_i \leq R}(R-d_i)} \sum_{i:d_i \leq R} (R-d_i)(\boldsymbol{p_i} - \boldsymbol{p})(\boldsymbol{p_i} - \boldsymbol{p})^T
	\end{equation}
	
	where $d_i = ||\boldsymbol{p_i} - \boldsymbol{p}||_2$ and $R$ is the spherical support region. As the computation of the covariance matrix assumes a spherical support region, we set the radius of the support region, $R$, as the farthest neighbour of the point $\boldsymbol{p}$, as given by Eq. \ref{eq:radius}. 
	
	\begin{equation}\label{eq:radius}
		R = \max_{1 \leq i \leq N} ||\boldsymbol{p_i} - \boldsymbol{p}||_2
	\end{equation}
	
	The region from the previous step is then aligned with this LRF for further processing. We denote this collection of points excluding the keypoint as $C$.
	
	\subsection{Descriptor Generation}
	For generating the descriptor, the projection of surface normals on each of the three axes, $x, y, z$ are computed. Let us denote the projection of a point $\boldsymbol{q}$ on axis $a$ where $a \in {\{x,y,z\}}$ as $q^{a}$. These are computed for the points in $C$ and are mathematically given by Eq. \ref{eq:projection}.

	\begin{equation}\label{eq:projection}
		q^x = \langle \boldsymbol{q}. \boldsymbol{\hat{i}} \rangle, \quad
		q^y = \langle \boldsymbol{q}. \boldsymbol{\hat{j}} \rangle, \quad
		q^z = \langle \boldsymbol{q}. \boldsymbol{\hat{k}} \rangle
		\quad \forall \ \boldsymbol{q} \in C 
	\end{equation}
	
	where $\langle.\rangle$ represents inner product and $\boldsymbol{\hat{i}, \hat{j}, \hat{k}}$ are the unit vector in the direction of x,y and z axes i.e. $(1,0,0)$, $(0,1,0)$ and $(0,0,1)$ respectively.
	
	Then the respective projections on each axis for pairs of ordered points $\boldsymbol{(q_m,q_n)}$ s.t. $1 \leq m,n \leq N$ are compared. The ordered pairs are constructed by first sorting the points in $C$ based upon the distance from the keypoint $\boldsymbol{p}$. 
	Let us denote the sorted set of points by $\boldsymbol{q_1}, \boldsymbol{q_2}...\boldsymbol{q_N}$ such that 
	
	\begin{equation} \label{eq:nndistance}
		||\boldsymbol{q_m} - \boldsymbol{p}||_2 \leq ||\boldsymbol{q_n} - \boldsymbol{p}||_2 \quad \forall \  m<n \  \text{and} \  m,n \in (1,N)
	\end{equation}
	Then the ordered set of points consist of the pairs 
	\begin{equation}
		(\boldsymbol{q_m}, \boldsymbol{q_n}) \ \text{s.t.} \ m<n 
	\end{equation}
	
	Each ordered pair thus obtained, results in three comparisons corresponding to projections on three axes. Each comparison is assigned a binary bit $b_a \in {\{0,1\}}$ for an axis $a$ based upon the relative magnitude of the projections, as given in Eq. \ref{eq:bitassign}.
	
	\begin{equation}\label{eq:bitassign}
		b_a=
		\begin{cases}
			1, & \text{if}\ q_m^a \geq q_n^q \\
			0, & \text{otherwise}
		\end{cases}
	\end{equation}
	
	The comparison for an ordered pair, $i$, results in a bit vector of size $3$ given by Eq. \ref{eq:orderedbit}.
	
	\begin{equation}\label{eq:orderedbit}
		\boldsymbol{b(i)} = b_xb_yb_z
	\end{equation}
	
	Finally, the descriptor is constructed by concatenating the binary strings $\boldsymbol{b(i)}$ from all the comparisons resulting in a binary vector of size $3*\frac{N*(N-1)}{2}$. The process is pictorially visualized as in Figure 2(b).
	
	
	
	\subsection{Descriptor Matching}\label{sec:descmatch}
	The descriptors are matched using Hamming Distance. There are efficient algorithms to compute hamming distances especially on CPUs having hardware support to count bits in a word (for ex: POPCNT instruction). In practical applications, there are usually millions of such descriptors, say, $D$,  which are needed to be matched. Performing a linear search over $D$ descriptors would be of the order of $O(D^{2})$, which would be computationally very expensive. For binary descriptors from 2D images, the popular way of matching these descriptors is to hash the binary bit vectors using techniques such as Locality Sensitive Hashing (LSH)\cite{rublee2011orb} and if needed, perform a linear search in each bucket. But, it is important to note here that the matching complexity of such techniques is affected by two important factors. Firstly, by the number of descriptors in the search space i.e. $D$ and secondly by the length of the binary bit vectors. Typically if the length of the binary vector is greater than $32$, various matching techniques use linear scan \cite{kulis2009learning}. These limitations are aggravated in the context of 3D point clouds as the number of points could be of the order of millions resulting in a huge number of keypoints. Moreover, the size of the proposed 3D Binary Signature is $O(N^2)$ for each keypoint, where $N$ is the number of nearest neighbour of a keypoint. Therefore, we adopt a fast binary feature matching technique proposed in \cite{muja2012fast}. The technique has been chosen for its low memory footprint and the ability to scale to large datasets. The technique is briefly described below.
	
	\subsubsection{Fast Feature Matching}\label{sec:fastmatching}
	
	\begin{itemize}
		\item \textit{Building search tree of  features}: The input data (descriptors) is divided into $K$ clusters by randomly selecting $K$ data points. The remaining data points are assigned to the closest cluster center (similar to k-medoids clustering). If the number of data points in a cluster ($DP_C$) is above a certain threshold i.e. maximum number of leaf nodes ($S_{L}$), then the algorithm is recursively repeated until each cluster has $DP_C < S_L$.
		\item \textit{Search for nearest neighbours}: The search is performed in parallel on multiple hierarchical clustering trees. The search begins by recursively exploring the node nearest to the query descriptor while the unexplored nodes are added to a priority queue. Once a node has been completely traversed, the next nearest node from the priority queue is extracted and is again explored recursively. This stopping criteria for this recursive search is based upon a search precision i.e. the fraction of exact neighbours discovered in the total number of returned neighbours.   
	\end{itemize}
	
	\section{Experiments and Results}\label{results}
	In this section, we describe the experimental details and the results thus produced.
	
	\subsection{Experimental Setup}
	\subsubsection{System Specification}
	The experiments were performed on a system having an Intel i7 processor with 128GB RAM and Ubuntu 14.04 Operating System. The implementation has been done in C++ (g++ 4.8) using Point Cloud Library (PCL) \cite{Rusu_ICRA2011_PCL} with OpenMP enabled.
	
	\subsubsection{Datasets}
	The experiments were performed on four publicly available datasets \cite{guo2014benchmark} with the details as shown in Table \ref{tab:dataset}. The chosen datasets allow us to evaluate the proposed descriptor on various application contexts. The \textit{Random Views, Laser Scanner} and \textit{LIDAR} datasets are suited for object recognition. In these datasets the model are full 3D meshes while the scenes are 2.5D views from specific viewpoints. On the other hand, the \textit{Retrieval} dataset is for 3D shape retrieval where the scenes are built by introducing rigid transformations and noise. Moreover, these datasets also allow for evaluating the descriptor on varying quality of point clouds. The \textit{Random Views} and \textit{Retrieval} have been derived from high resolution models. Comparatively, the \textit{Laser Scanner} and \textit{LIDAR} datasets can be categorized as medium and low quality respectively.
	
	\begin{table}[]
		\centering
		\caption{Datasets used in the study}
		\label{tab:dataset}
		\begin{tabular}{|l|r|r|}
			\hline
			\textbf{Dataset Name} & \multicolumn{1}{l|}{\textbf{\#Model}} & \multicolumn{1}{l|}{\textbf{\#Scene}} \\ \hline
			Random Views          & 6                                     & 36                                    \\ \hline
			Laser Scanner         & 5                                     & 10                                    \\ \hline
			LIDAR                 & 5                                     & 10                                    \\ \hline
			Retrieval             & 6                                     & 18                                    \\ \hline
		\end{tabular}
	\end{table}
	
	\subsection{Performance Evaluation}
	We evaluate the proposed descriptor for descriptiveness, compactness and efficiency against the best performing descriptors in the comparative analysis of \cite{guo2016comprehensive}. These descriptors are Fast Point Feature Histogram (FPFH) \cite{rusu2009fast}, Signature of Histogram of Orientations (SHOT) \cite{salti2014shot, tombari2010unique}, Unique Shape Context (USC) \cite{tombari2010unique} and Rotational Projection Statistics (RoPS) \cite{guo2013rotational}. As discussed in Section \ref{sec:kpdet},  Intrinsic Shape Signature (ISS) \cite{zhang2012microsoft} keypoint detector is used. We use the implementations available in Point Cloud Library (PCL) for the keypoint detector ISS and descriptors FPFH, SHOT, USC and RoPS.  Unless specified, the default parameters from the corresponding implementation are used for various techniques. The computation of best-fitting plane was performed in parallel on a GPU. For performing the fast feature matching (Section \ref{sec:fastmatching}), the number of parallel search trees has been fixed to $3$, branching factor to $16$ and maximum leaf nodes to $150$.
	
	\subsubsection{Descriptiveness}
	Descriptiveness is measured using Area under the \textit{Precision-Recall} curve (PRC). The PRC is generated using the following steps. Firstly, the keypoints are detected from the considered scenes and models. The keypoints are then described using various descriptors. For a fair comparison between the feature matching capability of the floating point and binary descriptors, we index the features using the technique described in Section \ref{sec:fastmatching} which can be applied consistently across floating point and binary descriptors. Due to this, the results of our experiments are slightly different from those reported in  \cite{guo2016comprehensive}. However, the relative performance results are still valid even though the absolute numbers change by a small margin (7.2\% on an average). The number of matches returned from the search tree depends upon the search precision $\tau$. A linear scan is performed on the matches obtained from the search tree and following \cite{tombari2013performance}, a match is considered correct if the matched features belong to the corresponding objects in the scene and model point clouds, and the matched keypoint lies within a small neighbourhood of the ground-truth. This neighbourhood is defined by a sphere of radius $2$ mesh resolution (mr) \cite{johnson1999using} and centered at the ground-truth keypoint. The \textit{precision} and \textit{recall} are then computed as 
	\begin{equation}
		Precision = \frac{\#Correct Matches}{\#Total Matches}
	\end{equation}
	
	\begin{equation}
		Recall = \frac{\#Correct Matches}{\#Corresponding Matches}
	\end{equation}
	
	We then vary $\tau$ from $0$ to $1$ and compute the Area under the PR curve ($AUC_{PR}$). The results are shown in Table \ref{tab:descriptive}. We denote the proposed binary descriptor with $N$ neighbours as 3DBS-N. The ranking for FPFH, USC, RoPS and SHOT are consistent with those reported in \cite{guo2016comprehensive}. It can be observed that 3DBS-32 outperforms other descriptors on  \textit{Retrieval} dataset while 3DBS-64 outperforms on all datasets except \textit{LIDAR} and \textit{Random Views}. Moreover, the magnitude of difference between $AUC_{PR}$ of RoPS and 3DBS with other descriptors is approximately $80\%$. This observation is important since it shows that the proposed technique has good performance on low resolution point cloud while also performing consistently for various transformations in other datasets. Another observation that can be made is that 3DBS-64 consistently performs better than 3DBS-32. This is expected since $64$ nearest neighbours span a larger local surface around a keypoint making the descriptor more robust to occlusion and clutter. This observation is in line with the performance of the other descriptors when an increase in the radius of the support region increases the performance of the descriptor \cite{guo2016comprehensive, tombari2013performance}
	
	\begin{table*}[]
		\centering
		\caption{Area under the Precision-recall curve ($AUC_{PR}$)}
		\label{tab:descriptive}
		\begin{tabular}{|l|r|r|r|r|r|r|}
			\hline
			\textbf{Dataset/Descriptor} & \textbf{FPFH} & \textbf{SHOT} & \textbf{USC} & \textbf{RoPS} & \textbf{3DBS-32} & \textbf{3DBS-64}    \\ \hline
			Random Views                & 0.24334       & \textbf{0.24799}       & 0.05982      & 0.20001       & 0.22341 & 0.24010 \\ \hline
			Laser Scanner               & 0.07341       & 0.05018       & 0.01103      & 0.16310       & 0.155953 & \textbf{0.16389}         \\ \hline
			LIDAR                       & 0.00198       & 0.00136       & 0.00164      & \textbf{0.00521}       & 0.004682   & 0.004897       \\ \hline
			Retrieval                   & 0.49319       & 0.56114       & 0.59521      & 0.52457       & 0.61109 & \textbf{0.64120} \\ \hline
		\end{tabular}
	\end{table*}

	\subsubsection{Compactness}
	Compactness of a descriptor is a measure to compare descriptors when memory footprint and storage requirements become important. Compactness is given as the 
	
	\begin{equation}
		Compactness = \frac{Average AUC_{PR}}{\#Floats_{descriptor}}
	\end{equation}
	
	The number of floats (length) in each of the considered descriptors is shown in Table \ref{tab:compact} with 32 bits per float. The results are graphically shown in Figure 3. It can be seen that 3DBS-32 is highly compact being close to FPFH. Moreover, 3DBS-64 has lower compactness than FPFH and is close to SHOT and RoPS. It would be important to note here that 3DBS can be made more compact by using bit vector compression schemes. Although the compactness measure is provided for completeness of comparative analysis with other popular 3D descriptors, it must be noted that the binary descriptors are not stored in memory as floats for computation. Therefore, compactness does not impact the matching efficiency of the proposed binary descriptor. 
	
	\begin{figure}[h]
		\caption{Compactness of Descriptors}
		\label{fig:compactness32}
		\centering
		\includegraphics[width=0.4\textwidth]{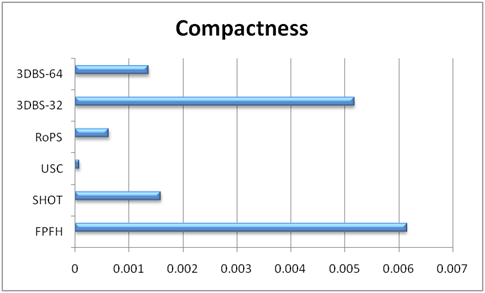}
	\end{figure}

	\begin{table}[]
		\centering
		\caption[Descriptor Length]{Descriptor Length}
		\label{tab:compact}
		\begin{tabular}{|l|r|}
			\hline
			\multicolumn{1}{|r|}{\textbf{Descriptor}} & \textbf{\# of floats} \\ \hline
			FPFH                                                              & 33                    \\ \hline
			SHOT                                                              & 135                   \\ \hline
			USC                                                               & 1980                  \\ \hline
			RoPS                                                              & 352                   \\ \hline
			3DBS-32                                                           & 48                    \\ \hline
			3DBS-64                                                           & 192                    \\ \hline
		\end{tabular}
	\end{table}
	
	\subsubsection{Efficiency}
\begin{figure}[h]
		\caption{Matching time comparison}
		\label{fig:time}
		\centering
		\includegraphics[width=0.4\textwidth]{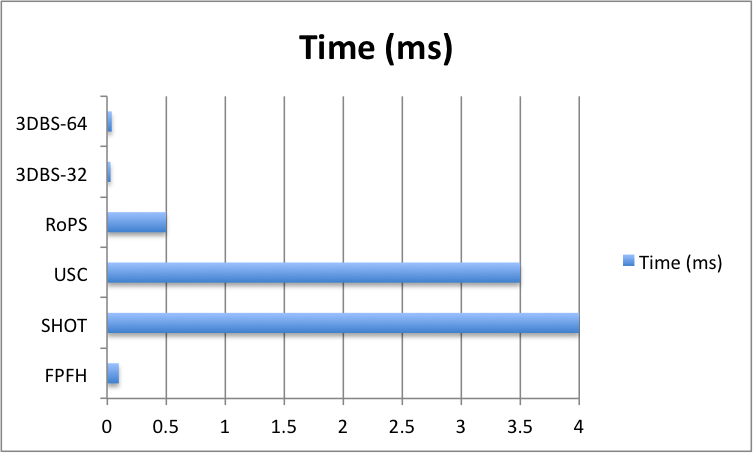}
	\end{figure}

	The major advantage of binary descriptors is that they can be matched extremely fast. To evaluate the descriptor matching time, the average matching time of keypoints on LIDAR dataset is reported in Figure \ref{fig:time}. It can be seen that 3DBS is nearly $10$ times as efficient than FPFH while almost three order of magnitude faster than other descriptors. This speed-up can be attributed to two factors. Firstly, the matching of binary vectors is by design faster than matching floating point vectors. Secondly, as discussed in Section \ref{sec:descmatch}, we leverage the built-in POPCNT instruction in GNU C Compiler providing tremendous computational efficiency in matching binary vectors. 
	
\begin{figure}[h]
		\caption{Performance on increasing NN}
		\label{fig:line}
		\centering
		\includegraphics[width=0.4\textwidth]{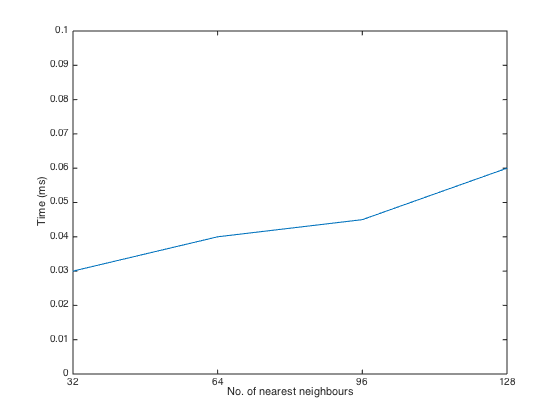}
	\end{figure}

	It can also be observed that the size of the proposed descriptor quadruples when the number of nearest neighbours doubles. As discussed previously, the number of nearest neighbours impacts the descriptiveness. Therefore, in Figure \ref{fig:line} we show the descriptor retrieval and matching time by gradually increasing the number of nearest neighbours. As can be seen, the increase in matching time is nearly sublinear when the number of nearest neighbours are doubled.
	
	\section{Conclusion}\label{conclusion}
	A novel 3D binary descriptor termed as 3D Binary Signature was proposed. The descriptor was based upon aligning the local surface as per a Local Reference Frame. The local surface has been identified with a nearest neigbour approach with angular constraint. This is in contrast with previous descriptors where a spherical region was used. The neighbourhood was characterized with projections of surface normals and encoding them as binary vector. We showed that the proposed descriptor outperforms the state of the art methods on various standard datasets. It is highly compact and nearly $3-10$ times faster than traditional descriptors, while demonstrating comparable or better descriptiveness. 
	
	\bibliographystyle{abbrv}
	\bibliography{sigproc}  
	%
	%

	
\end{document}